# A Topological derivative based image segmentation for sign language recognition system using isotropic filter

M.Krishnaveni,
Research Assistant,
Department of Computer Science,
Avinashilingam University for Women
Coimbatore, India

Dr.V.Radha,
Reader,
Department of Computer Science,
Avinashilingam University for Women
Coimbatore, India

*Abstract*-**The need of sign language is increasing radically especially to hearing impaired community. Only few research groups try to automatically recognize sign language from video, colored gloves and etc. Their approach requires a valid segmentation of the data that is used for training and of the data that is used to be recognized. Recognition of a sign language image sequence is challenging because of the variety of hand shapes and hand motions. Here, this paper proposes to apply a combination of image segmentation with restoration using topological derivatives for achieving high recognition accuracy. Image quality measures are conceded here to differentiate the methods both subjectively as well as objectively. Experiments show that the additional use of the restoration before segmenting the postures significantly improves the correct rate of hand detection, and that the discrete derivatives yields a high rate of discrimination between different static hand postures as well as between hand postures and the scene background. Eventually, the research is to contribute to the implementation of automated sign language recognition system mainly established for the welfare purpose.**

## I. INTRODUCTION

Currently, major research groups focus on the sign recognition problem. The recognition process includes segmentation as the chief step, but the segmentation results are not evaluated that good [18]. However, those recognition methods are based on several approaches that could also be used for sign segmentation [13]. The sign recognition methods can be classified into several categories according to the model of sign they refer to. The approaches are distinguished as one segment, multi segment and hidden model segment based. One-segment approach is used over this research work. In one-segment approach, each gesture is modeled as one single segment. This method has only been applied to gesture classification and was not employed to process real signs [2]. This kind of approach is mainly useful in sign language processing [14]. Here the research presents a segmentation approach to the automatic training and recognition of sign language. This work employs a enhancement method which uses filters with segmentation to locate the dominant hand. Both objective and subjective evaluation is measured by common parameters. The paper is organized as follows. In Section 2 we introduce the framework underlying the presented approach; Section 3 shortly introduces the restoration filters and Section 4 presents the segmentation method that is used in our approach. Section 5 presents the experimental results. Finally, the paper is summarized and concluded in section 7 with future work.

## II SYSTEM OVERVIEW

An overview of the automatic sign language recognition system is given here. This allows the research work to adopt the image processing techniques according to the need of the system.

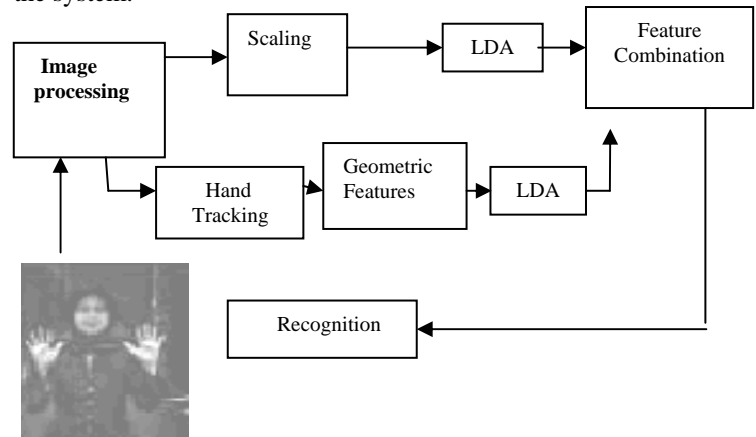

**Figure 1: System Overview**

There are many segmentation methods approved over sign language system. However these methods lacked precision due to the limited degree of freedom between the different hand signs [7]. This drawback resulted in common incorrect decisions since most hand signs are very similar thus not allowing the system to differentiate between them [16]. This research work over here is towards the enhancement strength before segmenting the hand sign which henceforth gives better performance in classifiers [8]. When the system runs the language model, transition and emission probabilities can be weighted by exponentiation with exponents α, β and γ, respectively, the probability of the knowledge sources are estimated as in eqn (1):

$$\Pr(w_1^N) \rightarrow p^\alpha(w_1^N),$$
$$\Pr(s_t \mid S_{t-1}, w_1^N) \rightarrow p^\beta(S_t \mid S_{t-1}, w_1^N), \text{----(1)}$$
$$\Pr(x_t \mid S_t, w_1^N) \rightarrow p^\gamma(x_t \mid S_t, W_1^N).$$

The exponents used for scaling α, β and γ are named language model scale, time distortion penalty, and word penalty, respectively. The system overview is shown in Figure. 1. The system overview is explained in such a way that the image is prescreened for hand tracking and from that the geometric features is extracted [9]. Linear discriminant analysis will be used for selecting combination of features that are to be trained through classifiers[10]. This paper will describe the intermediate part of how to segment the hand postures through topological derivatives combined with image restoration (filters).

### III. IMAGE RESTORATION TECHNIQUES

Noise occurs in all coherent imaging systems. To reduce the noise over the images two filters are used under continuum and discrete derivatives: isotropic and anisotropic filters. Here in this paper image restoration is handled with two filters which are shown in figure2. This approach before segmentation achieves the objective of detecting the number of open fingers using the concept of boundary tracing combined with finger tip detection [11]. It handles breaks, if any, during boundary tracing by rejoining the trace at an appropriate position. Here the restoration of the sign image is done using continuum and discrete topological derivative algorithm. The main idea behind this algorithm is to compute the topological derivative for an appropriate functional and a perturbation given by the introduction of cracks between pixels. This derivative is used as an indicator function to find the best pixels to introduce the cracks that, in the presence of diffusion, will most remove noise preserving relevant image characteristics[15]. Here, this paper shows the possibility to solve the image restoration problem using topological optimization. The basic idea is to adapt the topological gradient approach.

*A. Isotropic filtering* is a filter that enhance the given noisy image in particular point if it looks the same in all directions [4]. It reduces the noise cause as the non filters do. Isotropic filter is scalar filter F(x) : $IR^d \rightarrow IR$ is isotropic if it is invariant by rotations and symmetries ,i.e if eqn (2)

$$f(R^T x) = f(x) \forall R \varepsilon O(d) \text{----------------(2)}$$

A vector filter is isotropic eqn (3), if its kernel F(x) = F (x$_1$,…..x$_d$) : $IR^d \rightarrow M_d$ invariant by rotation and symmetries:

$$R.F(R^T x).R^t = F(x) \forall R \in O(d) \text{---------------(3)}$$

An Isotropic scalar filter f depends only on radial distance : f(x) = g (x$^T$x).This is not true anymore for isotropic vector filters.

Though the standard research proves that anisotropic filter is superior always, the scenario here states that the evaluation speaks positive only to isotropic filter, because

i) It works better for Gaussian noise.
ii) The image taken is two dimensional
iii) Defined only on unbounded domain.

*B. Anisotropic filtering* is a method of enhancing the image quality of textures[6] on surfaces that are at oblique viewing angles with respect to the camera where the projection of the texture (not the polygon or other primitive on which it is rendered) appears to be non-orthogonal[3]. Like bilinear and trilinear filtering it eliminates aliasing effects, but improves on these other techniques by reducing blur and preserving detail at extreme viewing angles.

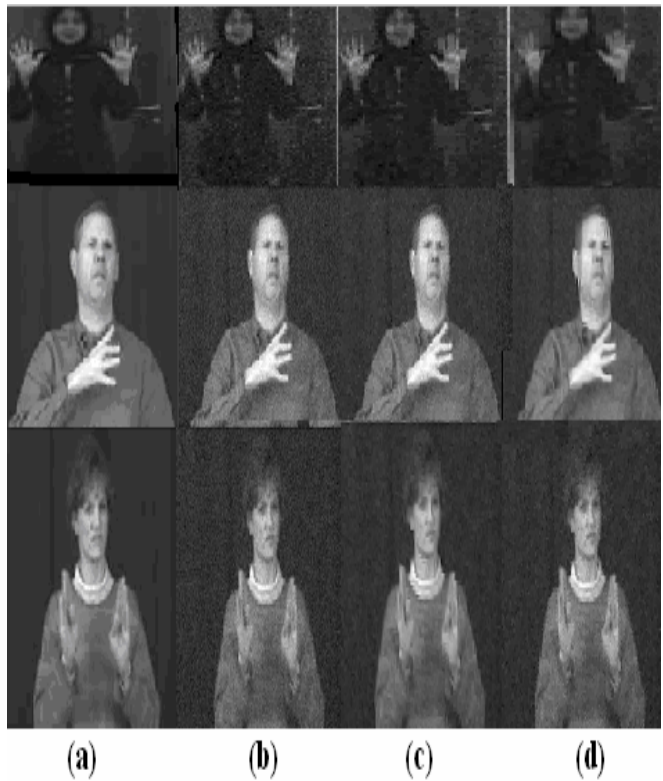

**Figure 2 Image restoration (a) Original Image (b) Noise (Gaussian) (c) isotropic filter (d) anisotropic filter**

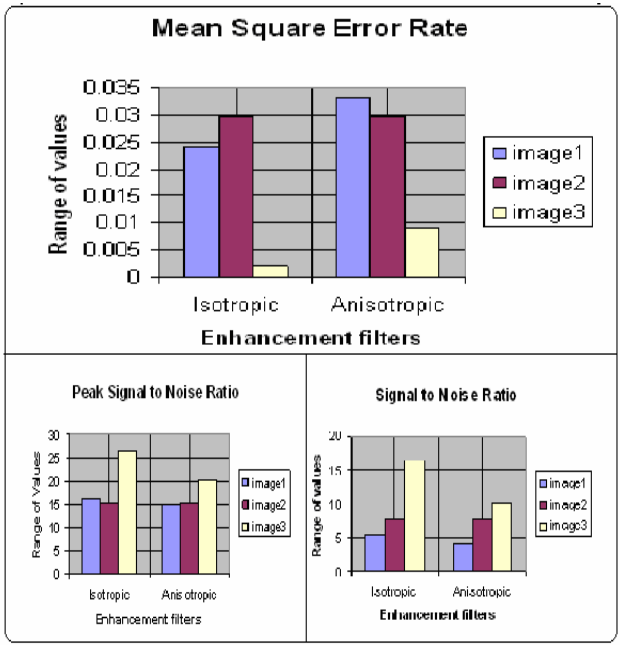

Figure 3: Performance evaluation of the images objectively

The isotropic filter performs better than the anisotropic filter in sign images. The reason for the anisotropic filter performing less optimally at low SNR is because the filter kernel is derived from the data, which itself is very noisy. As the SNR was increased, the isotropic filters were more optimal and were less prone to blurring and other over filtering effects. The study also demonstrates that with data at moderately high SNR, filtering can easily introduce more errors than are removed. In conclusion, the results from this study demonstrate that isotropic filtering can effectively reduce errors in sign images as long as they are applied properly.

## IV. IMAGE SEGMENTATION TECHNIQUES

There is no common solution to the segmentation problem in image processing domain. A priori knowledge about the objects present in the image, e.g., target, shadow, and background terrain, should be segmented in better way[14]. The visual relevance of the segmentation problems should be considered rather than simply their plurality; e.g. over-segmentation [5]. These techniques often have to be combined with preprocessing knowledge in order to effectively solve segmentation difficulties for a most needed application. One of the main purpose of proposed method is to precisely segment the image without the misplace of imperative information. Almost all image segmentation techniques are ad hoc in nature. Topological derivatives approach is used for image segmentation after best suited restoration process. Figure 3 demonstrates the experimental results of the segmentation work.

*A. Topological derivative*

Topological Derivative quantifies the sensitivity of a problem when the domain is perturbed by the introduction of heterogeneity (hole, inclusion, source term, etc.)[17][18]. Let the domain Ω under consideration is perturbed by the introduction of small holes (topology changes) in Ω as shown in fig4. Let us consider Ω be a bounded open set in $R^N$ (N=2,3) and $\gamma_\varepsilon$ be a crack of the ε centered at point xεΩ

$$\psi_\varepsilon(\Omega_\varepsilon) = \psi(\Omega) + f(\varepsilon)D_T(\hat{x}) + 0f((\varepsilon))\ldots\ldots\ldots\ldots(4)$$

Where f(ε) is known positive function going to zero with ε, and Dτ (x) is the topological derivative at point x given in eqn (4).

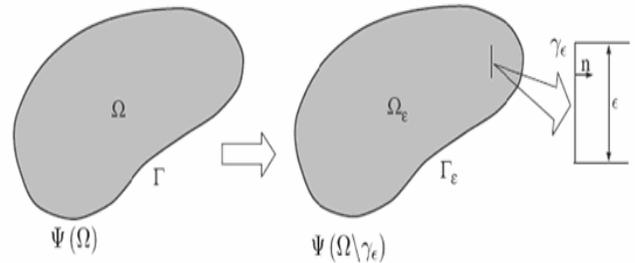

Figure 4: Concept of Topological Derivative

*a. Discrete topological derivative*

The algorithm is based on discrete topological derivative concept proposed by larrabide in which the cost functional used for discrete approach is represented by the eqn (5)

$$\Psi(\Omega_t^s) = \sum_s \sum_{pen} k^{s,p} \Delta\Omega_t^{s,p} . \Delta\Omega_t^{s,p} \ \text{----------(5)}$$

The Topological Derivative is given by the difference between perturbed cost function and original cost function.

*b. Continuum topological derivative*

Continuum (set theory), is known as the real line or the corresponding cardinal number. Continuum (theory), is nothing but, anything that goes through a gradual transition from one condition, to a different condition, without any abrupt changes. A single point (in the unique topology on a single point set) is a continuum ("is trivially a continuum", meaning that it satisfies the properties is easy and that the result is uninteresting); a continuum that contains more than one point. It is same as discrete but the design vector is b = {ϕ₁, ϕ₂.................ϕₙ}[18].These structure is modeled/analyzed as a continuum. Analysis models can therefore be large and expensive.

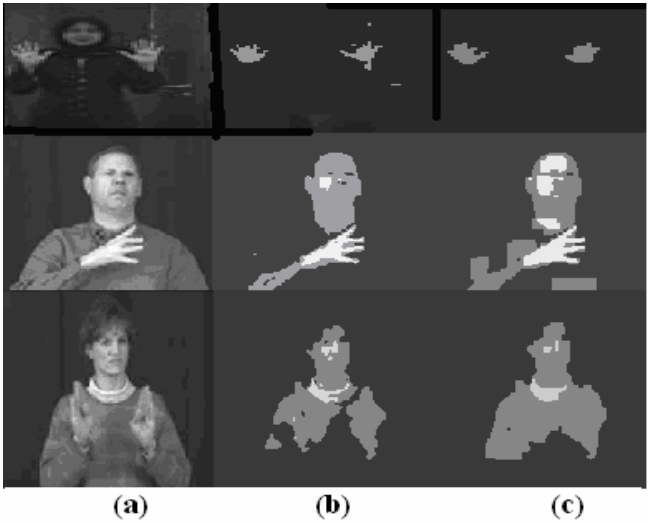

**Figure 5: Image segmentation using topological derivative algorithm (a) Original image (b) continuum derivative (c) Discrete derivative**

## V. PROPOSED WORK

Conventional problems of segmentation are known to be complicated and always in need of research as it implies mainly for object fragmentation. This research work is the expansion of enhancement and the need for the restoration filter before actual segmentation. Every joint framework outcome seems to be more prospective. This effort is validated with both subjective and objective experiments. The special significance of it is, it obtains promising results for isotropic filter.

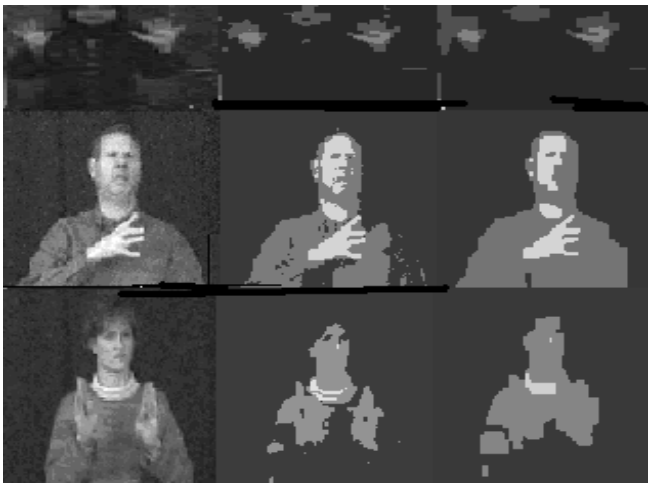

**Figure 6: Image segmentation using topological derivative algorithm after enhancement filtering of image (a) Original image (b) continuum derivative (c) Discrete derivative**

## VI PERFORMANCE EVALUATION

It is always vital when evaluating the performance of an algorithm when it is addressing a specific application. The relative performance of restoration and proposed method were evaluated and compared using the MSE criterion, PSNR difference and iteration. This reinforces the evaluation effort more suitably. The performance has its own uniqueness over the images taken. Figure 7 and 8 explain the objective evaluation of topological segmentation before and after filtering. Figure 9 demonstrates the performance evaluation of isotropic filter over topological derivative segmentation of sign images. It estimates the concert of segmentation methods reviewed for this work through iteration.

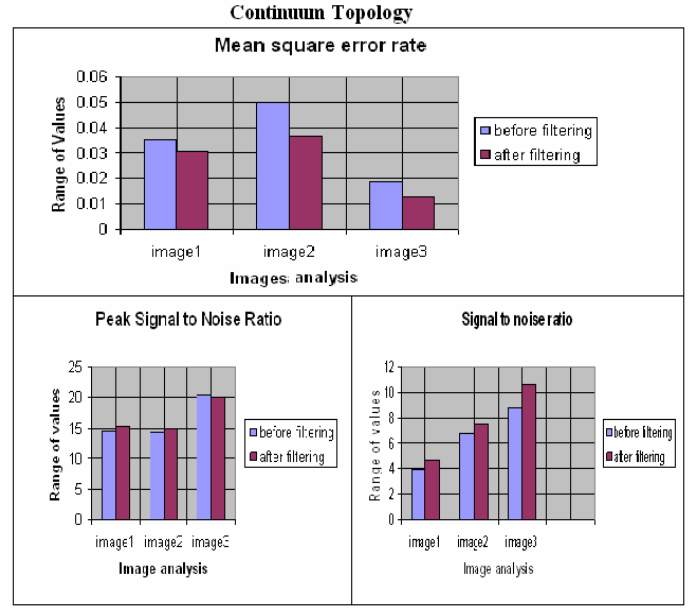

**Figure 8 Comparison of continuum topology before and after filtering**

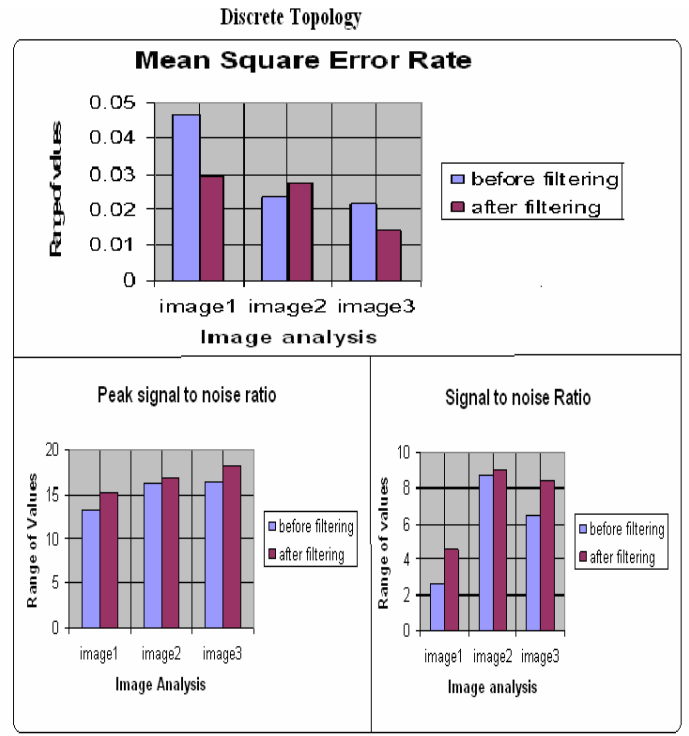

**Figure 8 Comparison of Discrete topology before and after filtering**

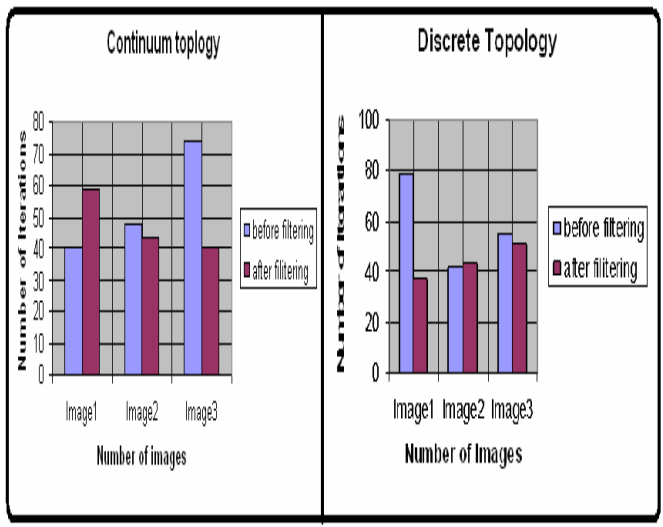

**Figure 9: Performance of the topology using the values of iterations**

## VII. CONCLUSION

The approach over here is to prove the significance of isotropic filter with topological derivative segmentation. The evaluation of image segmentation techniques is a key field of this study. The evaluation is categorized as objective and subjective. The result over the combination of isotropic filter with segmentation has given hopeful out come. The results indicate that the proposed approach is more robust and accurate than conventional segmentation methods mainly for foreground/ background segmentation evaluation problem. The computational time and number of iteration can be comparatively condensed by using optimization technique in further research.

## REFERENCES

[1] O. Friman, M. Borga, P. Lundberg, and H. Knutsson. Adaptive analysis of MRI data. *NeuroImage*, 19(3):837–845, 2003.

[2] O. Friman, J. Carlsson, P. Lundberg, M. Borga, and H. Knutsson. Detection of neural activity in functional MRI using canonical correlation analysis. *Magnetic Resonance in Medicine*, 45(2):323–330, February 2001.

[3] H. Knutsson, R. Wilson, and G. H. Granlund. Anisotropic non-stationary image estimation and its applications—Part I: Restoration of noisy images. *IEEE Transactions on Communications*, 31(3):388–397, March 1983.

[4] R. R. Nandy, C. G. Green, and D. Cordes. Canonical correlation analysis and modified ROC methods for fMRI techniques.In *Proceedings of the ISMRM Annual Meeting (ISMRM'02)*, Hawaii, USA, May 2002. ISMRM.

[5] Marroquín, J. L., Arce, E., and Botello, S., 2003, Hidden Markov Measure Field Models for Image Segmentation, *IEEE Transactions on Pattern Analysis and Machine Intelligence*, Vol. 25, No. 11, pp. 1380-1387.

[6]Orun, A. B., 2004, Automated Identification of Man-Made Textural Features on Satellite Imagery by Bayesian Networks. *Photogrametric Engineering & Remote Sensing*, Vol. 70, No. 2, pp. 211-216.

[7] J.J. Kuch and T.S. Huang, "Vision-based hand modeling and tracking for virtual teleconferencing and telecollaboration," in Proc. IEEE Int.Conf. Computer Vision, Cambridge, MA, pp. 666-671. June 1995.

[8] J. Davis and M. Shah, "Visual gesture recognition," Vision, Image, and Signal Processing, vol. 141, pp. 101-106, Apr. 1994.

[9] J. Rehg and T. Kanade, "DigitEyes: vision-based human hand tracking," School of Computer Science Technical Paper CMU-CS-93-220,Carnegie Mellon Univ., Dec.1993.

[10] Y. Shirai, N. Tanibata, N. Shimada, "Extraction of hand features forrecognition of sign language words," VI'2002,Computer-ControlledMechanical Systems, Graduate School of Engineering, Osaka University, 2002.

[11] C. Nölker, H. Ritter, "Detection of Fingertips in Human Hand Movement Sequences," Gesture and Sign Language inHuman-Computer Interaction, I. Wachsmuth and M. FroÈhlich, eds., pp.209-218, 1997.

[12] B. Bauer and H. Hienz, "Relevant features for video-based continuous sign language recognition," in Proc. of Fourth IEEE InternationalConference on Automatic Face and Gesture Recognition, pp. 440-445,March 2000.

[13] Y. Hamada, N. Shimada and Y. Shirai, "Hand Shape Estimation under Complex Backgrounds for Sign Language Recognition" , in Proc. of 6thInt. Conf. on Automatic Face and Gesture Recognition, pp. 589-594,May 2004.

[14] V. Mezaris,1,2 I. Kompatsiaris2 and M. G. Strintzis1,2 ,"Still Image Objective Segmentation Evaluation using Ground Truth", 1 Information Processing Laboratory, Electrical and computer Engineering Department,Aristotle University of Thessaloniki, Thessaloniki 54124, Greece 2 Informatics and Telematics Institute, 1st Km Thermi-Panorama Rd, Thessaloniki 57001, Greece, 5th COST 276 Workshop (2003), pp. 9–14.

[15] Auroux,D., Masmoudi,M., Belaid,L., 2006. 'Image restoration and classification by topological asymptotic expansion,' Variation Formulations in Mechanics: Theory and Applications- CIMNE, Barcelona, Spain (In press), (pp.1-16).

[16]Dirk Selle, Wolf Spindler, Bernhard Preim, and Heinz-Otto Peitgen., 2002. 'Mathematical Methods in Medical Imaging: Analysis of Vascular Structures for Liver Surgery Planning, (pp.1-21)

[17]Larrabid,I., Novotny,A.A., Feijo'o,R.A., and Taroco,E., 2005. 'A medical image enhancement algorithm based on topological derivative and anisotropic diffusion,' Proceedings of the XXVI Iberian Latin-American Congress on Computational Methods in Engineering CILAMCE, (pp1-14).

[18]Larrabide,I., Novotny,A.A., Feijo'o,R.A., Taraco,E., and Masmoudi,M., 2005 'An image segmentation method based on a discrete version of the topological derivative,' World Congress Structural and Multidisciplinary Optimization 6, Rio de Janeiro., International Society for Structural and Multidisciplinary Optimization, (pp.1-11).

**AUTHORS PROFILE**

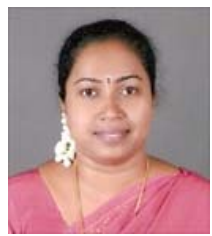

Ms. M. Krishnaveni, 3 Years of Research Experience Working as Research Assistant in Naval Research Board project, Area of Specialization: Image Processing, Pattern recognition, Neural Networks. Email id: krishnaveni.rd@gmail.com. She has 10 publications at national and International level journals and conferences.

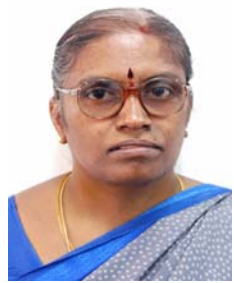

Dr. V. Radha, more than 20 years of teaching experience as Reader. Area of Specialization: Image Processing, Optimization Techniques, Voice Recognition and Synthesis. Email id: radharesearch@yahoo.com. She has 20 publications at national and international level journals and conferences.